\pdfoutput=1
\documentclass[11pt,a4paper]{article}
\usepackage[hyperref]{emnlp2018}
\usepackage{times}
\usepackage{latexsym}
\usepackage{graphicx}
\graphicspath{figure/}
\usepackage{url}
\usepackage{amsmath, amsfonts}
\usepackage{tikz}
\usepackage{tikz-qtree}
\usepackage{booktabs}
\usepackage{comment}

\aclfinalcopy 

\title{Evaluating Syntactic Properties of Seq2seq Output with a Broad Coverage HPSG: A Case Study on Machine Translation}

\author{Johnny Tian-Zheng Wei \\
  College of Natural Sciences \\
  University of Massachusetts Amherst \\
  \texttt{jwei@umass.edu} \\\And
  Khiem Pham \\
  Department of Computer Science \\
  San Jose State University \\
  \texttt{khiem.pham@sjsu.edu} \\\AND
  Brian Dillon \\
  Department of Linguistics \\
  University of Massachusetts Amherst \\
  \texttt{brian@linguist.umass.edu} \\\And
  Brendan O'Connor \\
  College of Information and Computer Sciences \\
  University of Massachusetts Amherst \\
  \texttt{brenocon@cs.umass.edu}
}

\date{8/30/2018}
\begin{document}
\maketitle
\begin{abstract}

Sequence to sequence (seq2seq) models are often employed in settings where the target output is natural language. However, the syntactic properties of the language generated from these models are not well understood. We explore whether such output belongs to a formal and realistic grammar, by employing the English Resource Grammar (ERG), a broad coverage, linguistically precise HPSG-based grammar of English. From a French to English parallel corpus, we analyze the parseability and grammatical constructions occurring in output from a seq2seq translation model.  Over 93\% of the model translations are parseable, suggesting that it learns to generate conforming to a grammar. The model has trouble learning the distribution of rarer syntactic rules, and we pinpoint several constructions that differentiate translations between the references and our model. \\
\end{abstract}

\section{Introduction}
Sequence to sequence models \cite[seq2seq;][]{ DBLP:conf/nips/SutskeverVL14, DBLP:journals/corr/BahdanauCB14} have found use cases in tasks such as machine translation \cite{DBLP:journals/corr/WuSCLNMKCGMKSJL16}, dialogue agents \cite{DBLP:journals/corr/VinyalsL15}, and summarization \cite{DBLP:conf/emnlp/RushCW15}, where the target output is natural language. However, the decoder side in these models is usually parameterized by gated variants of recurrent neural networks \cite{DBLP:journals/neco/HochreiterS97}, and are general models of sequential data not explicitly designed to generate conforming to the grammar of natural language.

The syntactic properties of seq2seq output is our central interest. We focus on machine translation as a case study, and situate our work among those of artificial language learning, where we train our translation model exclusively on sentence pairs where the target-side output is in our grammar, and test our models by evaluating the output with respect to a grammar. We attempt to understand seq2seq output with the English Resource Grammar \cite{DBLP:journals/nle/Flickinger00}, a broad coverage, linguistically precise HPSG-based grammar of English, and explore the advantages and potential of using such an approach.

\begin{figure}[t!]
\begin{center}
\begin{tabular}{r|l}
French & Une situation grotesque. \\
Reference & It is a grotesque situation. \\
NMT Output & A generic\_adj situation.
\end{tabular}

\begin{tabular}{r r}
\begin{tikzpicture}[scale=0.66]
\Tree [.root\_strict [.sb-hd\_mc [.hdn\_bnp-qnt It ] [.hd-cmp\_u is [.sp-hd\_n a  [.aj-hdn\_norm grotesque situation ] ] ] ] ]
\end{tikzpicture} & \begin{tikzpicture}[scale=0.66]
\Tree [.root\_frag [.np\_frg [.sp-hd\_n a [.aj-hdn\_norm generic\_adj situation ] ] ] ] ] ]
\end{tikzpicture}
\end{tabular}

\end{center}
\caption{\label{one_ex} A test set source-reference pair and the NMT translation. Below are parser derivations in the ERG of both the reference and NMT translation. The ERG is described in \S\ref{hpsg}. Non-syntactic rules have been omitted. The NMT model is trained and tested only on sentence pairs where the reference is parseable by the ERG. The NMT translation may not always be parseable. Analysis on model output parseability in \S\ref{parse}. }
\end{figure}

This approach has three appealing properties in evaluating seq2seq output.
First, the language of the ERG is a departure from studies on unrealistic artificial languages with regular or context-free grammars, which give exact analyses on grammars that bear little relation to human language \cite{DBLP:conf/icml/WeissGY18, DBLP:journals/tnn/GersS01}. In fact, about 85\% of the sentences found in Wikipedia are parseable by the ERG \cite{DBLP:conf/lrec/FlickingerOY10}. Second, our methodology directly evaluates sequences the model outputs in practice with greedy or beam search, in contrast to methods rescoring pre-generated contrastive pairs to test implicit model knowledge \cite{DBLP:journals/tacl/LinzenDG16, DBLP:journals/corr/Sennrich16}. Third, the linguistically precise nature of the ERG gives us detailed analyses of the linguistic constructions exhibited by reference translations and parseable seq2seq translations for comparison.

Figure \ref{one_ex} shows an example from our analysis. Each testing example records the reference derivation, the model translation, and the derivation of that translation, if applicable. The derivations richly annotate the rule types and the linguistic constructions present in the translations. 

Our analysis in \S\ref{parse} presents results on parseability by the ERG and summarizes its relation to surface level statistics using Pearson correlation. In \S\ref{grammaticality} we manually annotate a small sample of NMT output without ERG derivations for grammaticality. We find that 60\% of exhaustively unparseable NMT translations are ungrammatical by humans. We also identify that 18.3\% of the ungrammatical sentences could be corrected by fixing agreement attachment errors. We conduct a discriminatory analysis in \S\ref{discriminatory} on reference and NMT rule usage to guide a qualitative analysis on our NMT output. In analyzing specific samples, we find a general trend that our NMT model prefers to translate literally.

\section{Head-phrase Structure Grammars} \label{hpsg}

A head-phrase structure grammar \cite[HPSG;][]{PollardSag94} is a highly lexicalized constraint based linguistic formalism. Unlike statistical parsers, these grammars are hand-built from lexical entries and syntactic rules. The English Resource Grammar \cite{ DBLP:journals/nle/Flickinger00} is an HPSG-based grammar of English, with broad coverage of linguistic phenomena, around 35K unique lexical entries, and handling of unknown words with both generic part-of-speech conditioned lexical types \cite{DBLP:conf/lrec/AdolphsOCCFK08} and a comprehensive set of class based generic lexical entries captured by regular expressions. The syntactic rules give fine-grained labels to the linguistic constructions present.\footnote{A list of rules types and their descriptions can be found at \url{http://moin.delph-in.net/ErgRules}.} While the ERG produces both syntactic and semantic annotations, we focus only on syntactic derivations in this study.

Suitable to our task, the ERG was engineered to capture as many grammatical strings as possible, while correctly rejecting ungrammatical strings. Parseability under the ERG should have linguistic reality in grammaticality. Ideally, there will be no parses for any ungrammatical string, and at least one parse for all grammatical strings, which can be unpacked in order of scores assigned by the included maximum entropy model. We make a distinction between parseability and grammaticality. For our purposes of evaluating with a specified grammar, we consider the parseability of sentences under the ERG in \S\ref{parse}, regardless of human grammaticality judgments. In \S\ref{grammaticality}, we manually annotate unparseable sentences for English grammaticality.

All experiments are conducted with the 1214 version of the ERG, and the LKB/PET was used for all parsing \cite{DBLP:conf/lrec/CopestakeF00}. We use the default parsing configuration (command line option ``\texttt{--erg+tnt}''), which uses a parsing timeout of 60 seconds. A sentence is labeled unparseable either if the search space contains no derivations or if not a single derivation is found within the search space before the timeout. Figure \ref{one_ex} shows a simplified derivation tree.

\section{Experimental Setup} \label{exp_setup}

This section details our setup of a French to English (FR $\rightarrow$ EN) neural machine translation system which we now refer to as NMT. Our goal was to test a baseline system for comparable results to machine translation and seq2seq models.

\textbf{Dataset.} From 2M French to English sentence pairs in the Europarl v7 parallel corpora \cite{koehn2005europarl}, we subset 1.6M where the English/reference sentence was parseable by the ERG. For these 1.6M sentence pairs, we record the best tree of the English sentence as determined by the maximum entropy model included in the ERG. All sentence pairs we now consider have at least one English translation within our grammar, and we make no constraint on French. About 1.4M pairs were used for training, 5K for validation, and the remaining 200K reserved for analysis.

\textbf{Out of vocabulary tokens.} On the source-side French sentences, simple rare word handling was applied, where all tokens with a frequency rank over 40K were replaced with an ``UNK" token. However, when handling rare words in the target-side English sentences, ``UNK" will significantly degrade ERG parsing performance on model output. We replace our output tokens based on the lexical entries recognized by the ERG in our best parses (as in Figure \ref{one_ex}'s NMT output). This form of rare word handling is similar to the 10K PTB dataset \cite{DBLP:conf/interspeech/MikolovDKBC11}, but with more detailed part-of-speech and regular expression conditioned ``UNK" tokens. After preprocessing, we had a source vocabulary size of 40000, and a target vocabulary size of 36292.

\textbf{Model.}  Our translation model is a word-level neural machine translation system with an attention mechanism \cite{DBLP:journals/corr/BahdanauCB14}. We used an encoder and decoder with 512 dimensions and 2 layers each, and word embeddings of size 1024. Dropout rates of 0.3 on the source, target, and hidden layers were applied. A dropout of 0.4 was applied to the word embedding, which was tied for both input and output. The model was trained for about 20 hours with early stopping on validation perplexity with patience 10 on a single Nvidia GPU Titan X (Maxwell). We used the NEMATUS \cite{nematus} implementation, a highly ranked system in WMT16.

\begin{table}[t!]
\begin{center}
\begin{tabular}{r|r r|r r|r}
~ & \multicolumn{2}{c|}{\bf Strict} & \multicolumn{2}{c|}{\bf Informal} & \bf Unpar- \\
\bf Source & \bf Full & \bf Frag & \bf Full & \bf Frag & \bf seable \\
\hline
Ref & 64.7 & 2.4 & 31.5 & 1.4 & 0.0 \\
NMT & 60.5 & 3.0 & 28.1 & 1.6 & 6.8 \\
\hline
\bf $\Delta$ & -4.2 & +0.6 & -3.4 & +0.2 & +6.8 \\ 
\end{tabular}
\end{center}
\caption{\label{parseability} The distribution of root node conditions for the reference and NMT translations on the 200K analysis sentence pairs. Root node conditions are taken from the recorded best derivation. The best derivation is chosen by the maximum entropy model included in the ERG.}
\end{table}

\textbf{Translations.} After training convergence on the 1M sentence pairs, the saved model is used for translation on the 200K sentences pairs left for analysis. A beam size of 5 is used to search for the best translation under our NMT model. We parse these translations with the ERG and record the best tree under the maximum entropy model. We have parallel data of the French sentence, the human/reference English translation, the NMT English translation, the parse of the reference translation, and the parse of NMT translation (if it was grammatical). Note that the NMT translation may have no parse.

\section{Results} 

\subsection{Parseability} \label{parse}

\newcommand{\fracphantom}{\makebox[0pt][r]{\phantom{$\frac{\frac{P_x}{(P_x)}}{}$}}}
\begin{table}[t!]
\begin{center}
\begin{tabular}{r r | r }
\bf Feature & \bf Equation & $r$ \\
\hline
LP NMT & \fracphantom $\log P_m(S_o)$ & 0.313 \\
LP Unigr. (src-fr) & \fracphantom $\log P_{u}(S_i)$ & 0.289 \\
LP Unigr. (ref-en) & \fracphantom $\log P_{u}(S_r)$ & 0.273 \\
LP Unigr. (out-en) & \fracphantom $\log P_{u}(S_o)$ & 0.304 \\
Length Output & \fracphantom $|S_o|$ & -0.320 \\
Mean LP & \fracphantom $\frac{\log P_m(S_o)}{|S_o|}$ & 0.093 \\
Norm LP & \fracphantom $-\frac{\log P_m(S_o)}{\log P_{u}(S_o)}$ & 0.057 \\
\hline
\end{tabular}
\end{center}
\caption{\label{kendall} Pearson's $r$ of surface statistics against the binary parseability variable. Parseable is denoted with +1. $S_i, S_r, S_o$ are the input, reference, and NMT output sentences, respectively. We abbreviate log probability as ``LP.'' $P_m(S)$ is the probability of $S$ occurring under the NMT model, and $P_u(S)$ is the probability of $S$ occurring under a unigram model. }
\end{table}


The NMT translations for the 200K test split were parsed. Parsing a sentence with the ERG yields one of four cases:
\begin{itemize}
    \item Parseable. A derivation is found and recorded by the parser before the timeout. The best derivation is chosen by the included maximum entropy in the ERG. About 93.2\% of the sentences were parseable.
    \item Unparseable due to resource limitations. The parser reached its limit of either memory or time before finding a derivation. This constitutes about 3.2\% of all cases, and 47\% of unparsable cases. 
    \item Unparseable due to parser error. The parser encountered an error in retrieving lexical entries or instantiating the parsing chart. This constitutes about 0.5\% of all cases, and 8\% of unparsable cases.
    \item Unparseable due to exhaustation of search space. The parser exhausted the entire search space of derivations for a sentence, and concludes that it does not have a derivation in the ERG. This constitutes about 3.1\% of all cases, and 45\% of unparsable cases. 
\end{itemize}

\noindent
The distribution of the root node conditions for the reference and NMT translation derivations are listed in table \ref{parseability}, along with the parseability of the NMT translations. Root node conditions are used by the ERG to denote whether the parser had to relax punctuation and capitalization rules, with ``strict'' and ``informal'', and whether the derivation is of a full sentence or a fragment, with ``full'' and ``frag''. Fragments can be isolated noun, verb, or prepositional phrases. Both full sentence root node conditions saw a decrease in usage, with the strict full root condition having the largest drop out of all conditions. Both fragments have a small increase in usage.

\begin{figure}
\centering
\includegraphics[scale=0.38]{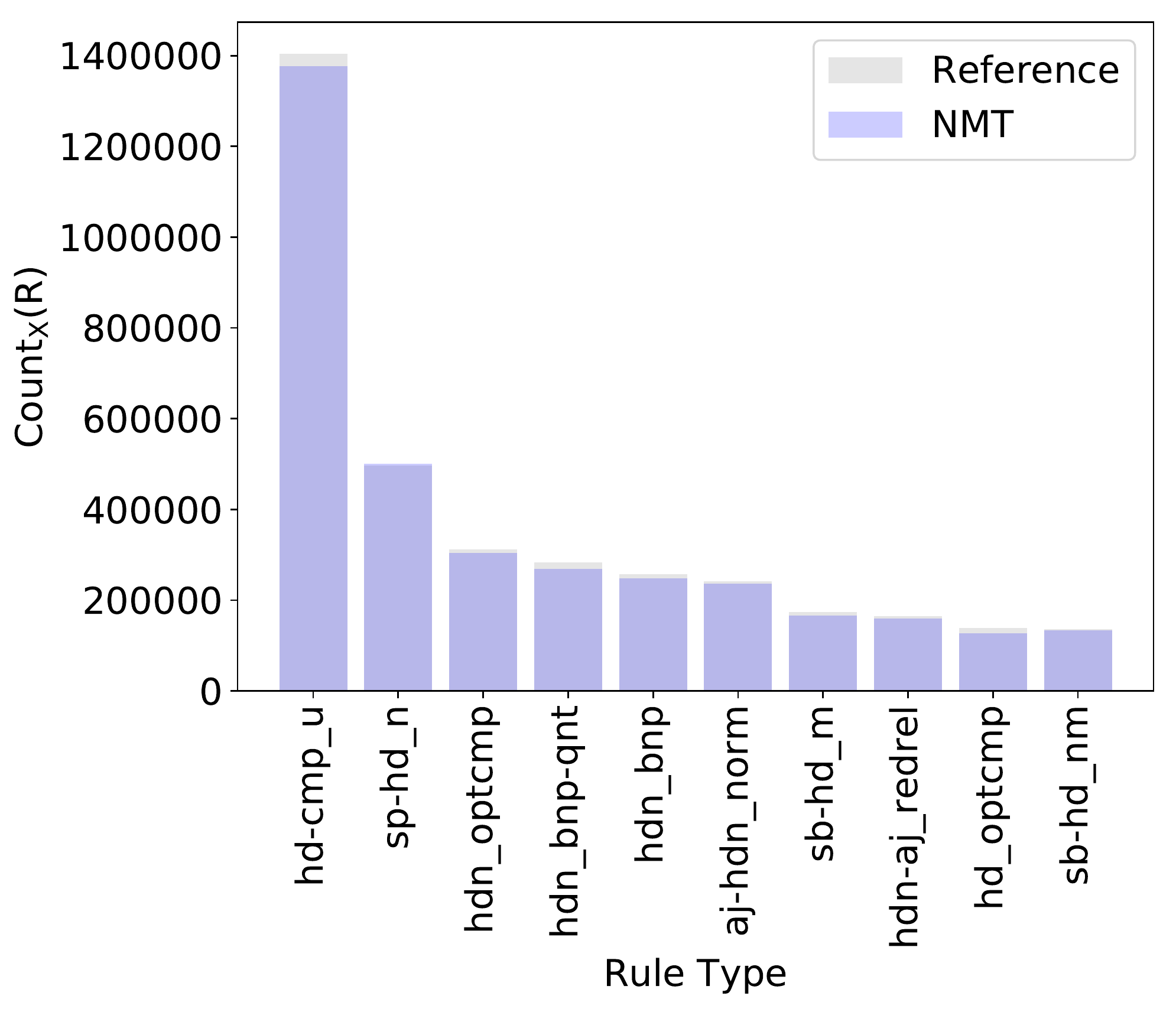}
\caption{\label{rule_dist} Count of rule usage for the 10 most frequent rules in the derivations of the reference and grammatical NMT translations.}
\end{figure}

We summarize the parseability of NMT translations with a few surface level statistics. In addition to log probabilities from our translation model, we provide several transformations of these scores, which were inspired by work in unsupervised acceptability judgments \cite{DBLP:conf/acl/LauCL15}. In table \ref{kendall}, we calculate Pearson's $r$ for each statistic and the binary parseability variable. The $r$ coefficient is effectively a normalized difference in means.

From the correlation coefficients, we see that the probabilities from the NMT and unigram models are all indicative of parseability. The higher the probabilities, the more likely the translation is to be grammatical. Length is the only exception with a negative coefficient, where the longer a sentence is, the less likely a translation is grammatical. Length has the strongest correlation of all our features, but this correlation may be due to 
limitations in the ERG's ability to parse longer sentences, instead of the NMT model's to generate longer grammatical sentences. We see that the LP NMT has a higher correlation with grammaticality than the unigram models, but not by a large amount. Coefficients for length and LP NMT have the two greatest magnitudes.

\subsection{Grammaticality} \label{grammaticality}

\begin{figure}
\centering
\includegraphics[scale=0.38]{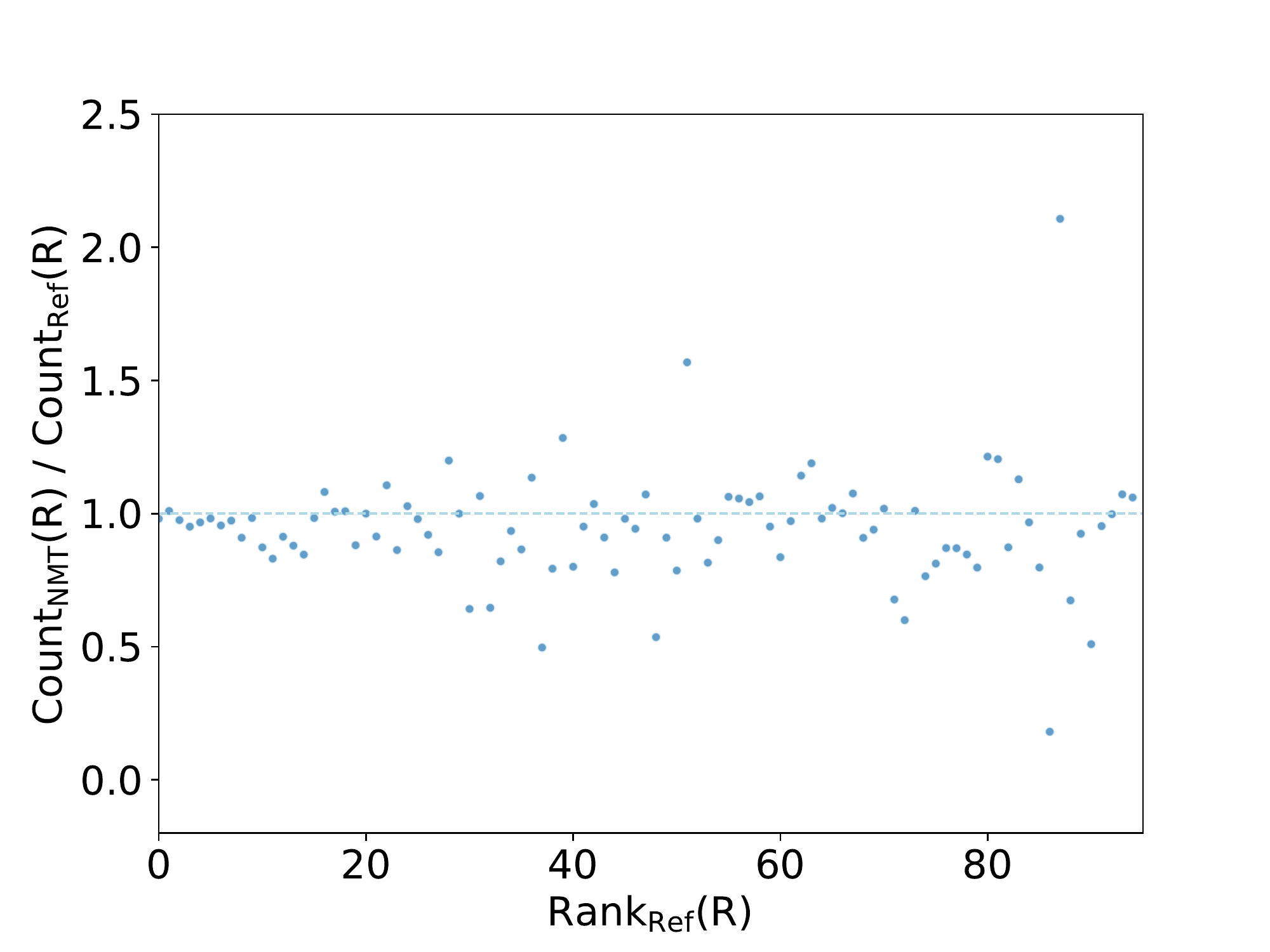}
\caption{\label{rank_per_diff} The ratio of each rule's count in grammatical NMT translations over count in reference translations, ordered by the rule's frequency rank in reference derivations. Only rules with over 1000 usages in the set of reference derivations are shown. }
\end{figure}

Out of the 14K unparseable NMT translations, there are 6.2K translations where the parser concluded unparseability after exhausting the search space for derivations. We will refer to these examples as ``exhaustively unparseable.'' To understand the relation between English grammaticality and exhaustive unparseability under the ERG, two linguistics undergraduates (including the first author) labeled a random sample of 100 NMT translations from this subset. We sampled only those translations with less than 10 words to limit annotator confusion. Annotators were instructed to assign a binary grammatical judgment to each sentence, ignoring the coherence and meaning of the translation, to the best of their abilities. Punctuation was ignored in all annotations, although the ERG is sensitive to punctuation. When the sentence was ungrammatical, subject-verb agreement and noun phrase agreement errors were annotated.

\begin{table*}[t!]
\begin{center}
\begin{tabular}{|r r|r r|}
\hline
\multicolumn{2}{|c|}{\bf Reference} & \multicolumn{2}{c|}{\bf NMT} \\
\bf Rule Type & \bf Annotations & \bf Rule Type & \bf Annotations \\
\hline
\hline
xp\_brck-pr & Paired bracketed phrase & j\_sbrd-pre & Pred.subord phr fr.adj, prehead \\
cl-cl\_runon & Run-on sentence w/two clauses & n-j\_j-cpd & Compound from noun+adj \\
np-hdn\_cpd & Compound proper-name+noun & j\_n-ed & Adj-phr from adj + noun+ed \\
vp\_sbrd-prd-prp & 	
Pred.subord phr from prp-VP & aj-np\_int-frg & 	
Fragment intersctv modif + NP \\
hd-aj\_int-sl & Hd+foll.int.adjct, gap in adj & vp\_sbrd-prd-aj & Pred.subord phr from adjctv phr \\
hd-aj\_vmod & 	
Hd+foll.int.adjct, prec. NP cmp & np\_frg & Fragment NP \\
vp\_np-ger & NP from verbal gerund & flr-hd\_nwh & Filler-head, non-wh filler \\
mrk-nh\_atom & Paired marker + phrase & hdn-aj\_rc-pr & NomHd+foll.rel.cl, paired pnct \\
vp\_sbrd-pre & Pred.subord phr fr.VP, prehead & sb-hd\_mc & Head+subject, main clause \\
num\_prt-det-nc & Partitive NP fr.number, no cmp & num-n\_mnp & Measure NP from number+noun \\
\hline
\end{tabular}
\end{center}
\caption{\label{lr-discrim} The most discriminatory features of both the reference and NMT translations. Features are ranked by a logistic regression without an intercept and an L1 penalty $C=0.01$, trained with LIBLINEAR within scikit-learn. Description of rule types are taken from the annotations in the ErgRules website.}
\end{table*}

Within our random sample, 60 sentences were labeled as ungrammatical. Of these ungrammatical sentences, 5 could be made grammatical if a subject-verb agreement error was corrected, and 5 other translations could be made grammatical by correcting an article or determiner attachment to a noun. One translation exhibited both forms of agreement attachment errors. Agreement attachment errors are better studied phenomenon \cite{DBLP:journals/tacl/LinzenDG16, DBLP:journals/corr/Sennrich16}. However, correcting these errors only fixes 18.3\% of ungrammaticality that we observed in our sample.

Out of the 100 sampled NMT translations that have no ERG derivations, we found 35 to be grammatical. 5 test examples were excluded. These include two cases where the source sentences were empty, and three cases where the sentence was parliament session information. Both annotators found annotating to be challenging, and possibly better annotated on an ordinal scale. Out of the exhaustively unparseable random sample, 37\% was found to be grammatical. The ERG may have grammar gaps for near grammatical sentences.

\subsection{Rule Counts}

This section and those following will analyze the rules present in the derivations of the reference and the grammatical NMT translations. We consider only the appearance of the rule, disregarding the context it appears in, and define $\text{Count}_\text{X}(R)$ as the number of times rule $R$ appears in the set $X \in \{ \text{Ref}, \text{NMT} \}$ of derivations. In figure \ref{rule_dist}, we plot the counts of the 10 most frequent rule types in the reference and NMT translations. The rules were taken from the best derivations as determined by the included maximum entropy classifier in the ERG. Note that we have about 200K reference derivations and 189K NMT derivations we aggregate statistics from, as about 7\% of the NMT translations are unparseable. We see that both distributions seem to be Zipfian, and that the rule counts in the NMT translations match the reference closely. 

In figure \ref{rank_per_diff}, for each rule $R$, we plot the ratio $\text{Count}_{\text{NMT}}(R) / \text{Count}_{\text{Ref}}(R)$ of derivations against the rank of the rule type. The rank is computed from the set of reference derivations. The variance of the ratio seems to increase as the rank of the rule increases. While the occurrences of rarer constructions is low in the NMT translations,  it seems not to match the  usage in the reference translation dataset. This suggests that NMT has trouble learning the usage of rarer syntactic constructions.

\subsection{Discriminative Rules} \label{discriminatory}

This section aims to understand which usage of rules distinguish the reference from the NMT translations. The analysis in this section is largely inspired by work in syntactic stylometrics \cite{DBLP:conf/emnlp/FengBC12, DBLP:conf/emnlp/AshokFC13}, where we vectorize each derivation as a bag of rules, and fit a logistic regression without an intercept to predict whether a derivation was from the set of reference or NMT translations. In total, there are 392K examples and we prepare an 80/20 training validation split. The model is fit with an L1 sparsity penalty of $C=0.01$ with the LIBLINEAR solver in scikit learn \cite{scikit-learn}. On the validation set, the logistic regression achieves an accuracy of about 59.0\% on the validation set up from the 51.9\% majority class baseline. Of the 204 rules used as features, only 71 were non-zero. There are 47 rules that are discriminatory towards reference translations (positive weights), and 24 rules that are discriminatory towards NMT translations (negative weights). Table \ref{lr-discrim} shows the 10 most discriminative rules for each set.

\subsection{Qualitative Analysis}
We provide qualitative analysis for a few of the most discriminative rules for both the reference and NMT translations. When exploring discriminatory rules in the reference, we sampled for sentence pairs where the reference translation that contained the rule of interest, and the NMT translation did not. We only sampled within sentences with a length of less than 12. Our qualitative analysis is written after we looked through many samples, and we attempted to list a few of our general observations for each rule.

The ``cl-cl\_runon'' rule type indicates a runon sentence with two conjoined clauses. This rule has a positive coefficient, and discriminates towards reference translations. An example is given below:
\begin{figure}[!htb]
\begin{center}
\begin{tabular}{r|l}
French & je le r\'ep\`ete , vous avez raison . \\
Reference & i repeat ; you are quite right . \\
NMT Output & i repeat , you are right .
\end{tabular}
\end{center}
\end{figure}

\noindent In this case, the NMT used a comma to conjoin two clauses instead of using a semi-colon, which is more similar in punctuation to the source sentence. In every case we saw, the NMT model seems to follow the French style of conjunction more closely, mirroring the punctuation of the source sentence. Reference translations seem to be more spurious in the usage of semicolons or periods. In more concerning cases, short conjoined clauses were dropped by the NMT translations; e.g. ``thank you .". 

We now analyze ``np\_frg'' which denotes a noun phrase fragment. This rule that has a negative coefficient, and discriminates towards NMT translations. We give an example below:
\begin{figure}[!htb]
\begin{center}
\begin{tabular}{r|p{5cm}}
French & quel paradoxe ! \\
Reference & what a paradox this is ! \\
NMT Output & what a paradox !
\end{tabular}
\end{center}
\end{figure}

\noindent When looking through samples, we saw many examples where the expletive is dropped. This is similar to the case for the previous rule as it is a literal translation of the French source. In NMT translations we observed increases in the formal and strict fragment root conditions, and we believe these translations are a factor.

\section{Related Work}
Previous work in recurrent neural network based recognizers on artificial languages has studied the performance on context-free and limited context-sensitive languages \cite{DBLP:journals/tnn/GersS01}. More recent research in this setting provide methods to extract the exact deterministic finite automaton represented by the RNN based recognizers of regular languages \cite{DBLP:conf/icml/WeissGY18}. These studies give exact analyses of RNN recognizers for simple artificial languages.

In the evaluation of language models in natural language settings, recent work analyzes the rescoring of grammatical and ungrammatical sentence pairs based on specific linguistic phenomenon such as agreement attraction \cite{DBLP:journals/tacl/LinzenDG16}. These contrastive pairs have also found use in evaluating seq2seq models through rescoring with the decoder side of neural machine translation systems \cite{DBLP:journals/corr/Sennrich16}. Both studies on contrastive pairs evaluate implicit grammatical knowledge of a language model. 

HPSG-based grammars have found use in evaluating human produced language. To determine the degree of syntactic noisiness in social media text, parseability under the ERG was examined for newspaper and Twitter texts \cite{DBLP:conf/ijcnlp/BaldwinCLMW13}. In predicting grammaticality of L2 language learners with linear models, the parseability of sentences with the ERG was found to be a useful feature \cite{DBLP:conf/acl/HeilmanCMLMT14}. These studies suggest parseability in the ERG has some degree of linguistic reality.

Our work combines analysis of neural seq2seq models with an HPSG-based grammar, which begins to let us understand the syntactic properties in the model output. Recent work most similar to ours is in evaluating multimodel deep learning models with the ERG \cite{DBLP:journals/corr/KuhnleC17a}. While their work uses the ERG for language generation to test language understanding, we evaluate language generation with the parsing capabilities of the ERG, and study the syntactic properties.

\section{Conclusion}

Neural sequence to sequence models do not have any explicit biases towards inducing underlying grammars, yet was able to generate sentences conforming to an English-like grammar at a high rate. We investigated parseability and differences in syntactic rule usage for this neural seq2seq model, and these two analyses were made possible by the English Resource Grammar. Future work will involve using human ratings and machine translation quality estimation datasets to understand which syntactic biases are preferable for machine translation systems. The ERG also produces Minimal Recursion Semantics \cite[MRS;][]{Copestake2005}, a semantic representation which our work does not yet explore. By matching the semantic forms produced, we can make evaluations of language generation systems on a semantic level as well. In using these deep resources for evaluation, there is a shortcoming in the biased coverage of the grammar. Future work will also study how to evaluate our models despite these limitations. We hope this paper spurs others' interest in HPSG-based or language-like grammar evaluations of neural networks. 

\section*{Acknowledgments}
This work was facilitated by the Center for Study of Language and Information REU site at Stanford University, where the first and second author were roommates. Andrew McCallum and IESL provided computing resources. Dan Flickinger, Chris Potts, and the anonymous reviewers provided helpful comments. Nicholas Thomlin co-annotated the grammaticality judgments. Technical support on DELPH-IN was provided by Michael Goodman and Stephen Oepen. The first author received mentorship from Omer Levy, Roy Schwarz, Chenhao Tan, Nelson Liu, and Noah Smith in the summer of 2017, and from Ari Kobren, Nicholas Monath, and Haw-Shiuan Chang in the years prior. The second author received mentorship from Guangliang Chen.

\bibliography{emnlp2018.bbl}
\bibliographystyle{acl_natbib_nourl}

\end{document}